\documentclass{article}
\usepackage{spconf,amsmath,graphicx}
\usepackage{graphicx}
\usepackage{amsmath}
\usepackage{amssymb}
\usepackage{booktabs}
\usepackage{subcaption}
\usepackage{graphicx}
\usepackage{caption}
\usepackage{mathtools}
\usepackage{amsmath,bm}
\usepackage{tablefootnote}
\usepackage{times}
\usepackage{bm}

\title{AN IMPARTIAL TRANSFORMER FOR STORY VISUALIZATION}
\name{Nikolaos Tsakas \qquad Maria Lymperaiou \qquad Giorgos Filandrianos \qquad Giorgos Stamou}
\address{National Technical University of Athens}
\begin{document}
%\ninept
%
\maketitle
\begin{abstract}
Story Visualization is an advanced task of computed vision that targets sequential image synthesis, where the generated samples need to be realistic, faithful to their conditioning and sequentially consistent. Our work proposes a novel architectural and training approach: the Impartial Transformer achieves both text-relevant plausible scenes and sequential consistency utilizing as few trainable parameters as possible. This enhancement is even able to handle synthesis of 'hard' samples with occluded objects, achieving improved evaluation metrics comparing to past approaches.
\end{abstract}
\begin{keywords}
Story Visualization, GANs, Transformers
\end{keywords}
\section{Introduction}
\label{sec:intro}
The emergence of GANs \cite{goodfellow2014gans} has inspired several advancements in image synthesis, one of the most prominent being conditional image synthesis with the usage of cGANs \cite{odena2017conditional}. Text-conditioned image generation has been a popular variant of the conditional case, displaying a long line of architectural exploration. Those topics stimulated the novel task of Story Visualization (SV), where a visual story needs to be generated conditioned on text or other semantic information. The
images need not only to correspond to their conditioning,
but also to remain consistent within the sequence, which requires a global understanding of the story context. The basic idea involves a GAN-based variant with one generator $G$ and two discriminators. The first discriminator (image discriminator $D_{im}$) focuses on text-image relevance, while the other one (story discriminator $D_{st}$) ensures the overall sequential coherence. The same task can be viewed as a sequence transduction problem, a task widely explored with the usage of recurrent neural networks (RNNs) and Transformers \cite{transformer}.

So far, SV has only received a few improvements, while it faces  scarcity of viable datasets and evaluation methods. To this end, we propose a refined transformer-based approach, where a simple and lightweight adjustment called \textbf{\textit{Impartial transformer}}  is enough to resolve problems present in our predecessors. A transformer encoder jointly trained from $G$ and $D_{im}$ is employed to create an input representation, yielding a resource-friendly scenario comparing to using separate encoders for each generative component or adding a plethora of modules \cite{Maharana2021ImprovingGA, Maharana2021IntegratingVL} to achieve advanced results

\section{Related work}
\label{sec:literature}
Generative Adversarial Networks (GANs) \cite{goodfellow2014gans} are able to synthesize high-quality images by initially receiving random noise $z \sim  p_z$ in the input of $G$ and are trained to gradually improve the synthesized sample from receiving feedback regarding sample quality from $D$. Conditional GANs (cGANS) also receive a conditioning vector $y$ among with $z$ to guide synthesis towards certain areas of the target distribution. Earlier works in conditional synthesis where $y$ is in textual form attempt to fully synthesize the final image in one step, resulting in samples lacking in fidelity \cite{reed2016t2i}. The first significant improvements emerged with the introduction of StackGAN \cite{stackgan} and its variants \cite{zhang2018stackgan} which gradually upsample images up to the final resolution. Further implementations target detail refinement \cite{xu2018attngan, zhu2019dmgan} and improvements of text-image relevance \cite{segan}. Proceeding to the sequential case, StoryGAN \cite{storygan} introduced the SV task utilizing RNNs for conditional encoding, as well as the two-discriminator GAN architecture that later variants follow \cite{pororogan, li2020storygan}. Only recently transformer-based approaches for conditional encoding emerged \cite{Maharana2021ImprovingGA, Maharana2021IntegratingVL} indicating a new direction of research obeying to recent trends \cite{transformer}.

\section{Method}
\label{sec:method}
\label{sec:architecture}
We propose an updated framework for the SV task based on the emergence of transformer-based techniques for sequence processing. Primarily, we recommend the use of a transformer encoder \cite{transformer} as a replacement for the RNN structure of StoryGAN \cite{storygan}, focusing on its optimal training regime. 
%Stepping also on the simplicity and efficiency of StackGAN \cite{stackgan} we extend the original idea by experimenting  with three attention mechanisms.

\begin{figure}[b]
\centering
\includegraphics[width=8cm]{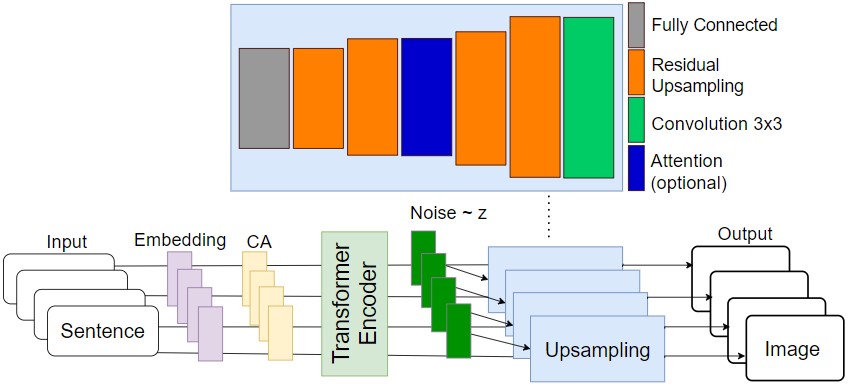}
\caption{The generator $G$ network (T = 4 frames)}
\label{fig:gen}
\end{figure}

\subsection{Generator}
The input to the generator $G$ is a sequence of symbols $s_t$, embedded by an encoder into vector representations $\phi_t$, t $\in$ [1, \textit{T}] where  \textit{T} corresponds to the length of all stories. Fig. \ref{fig:gen} depicts the basic $G$ architecture.

We recommend using a conditioning augmentation (CA) module, similar to \cite{stackgan}: Instead of conditioning the GAN on an embedding of the input $\phi_t$, a random vector $\hat{c}$ is sampled from a Gaussian distribution $\mathcal{N}(\mu(\phi_t,\Sigma(\phi_t)))$ with the mean $\mu(\phi_t)$ and the diagonal covariance matrix $\Sigma (\phi_t)$ being functions of the input embeddings. 
The vector $\hat{c}$ serves as the conditioning variable. CA promotes continuity in the data manifold, and can be also used to map the dimension of $\phi_{t}$ to its appropriate size. Training the parameters of this stochastic process becomes possible using the reparametrization trick \cite{kingma2014autoencoding}, where a sample 
from a Gaussian distribution with arbitrary mean $\mu$ 
and covariance matrix $\sigma$  can be produced as: $\hat{c} = \mu + z * \sigma$, where $z\sim\mathcal{N} (0,1)$. In addition, 
to ensure the smoothness of the manifold, the KL divergence between the 
learned Gaussian distribution and the standard one is added to the loss function of $G$ as a regularization term, therefore avoiding overfitting caused by collapsing to a  single point or by a distribution that deviates from the standard Gaussian \cite{stackgan}:

\begin{center}
    $Loss_{KL} = D_{KL}(\mathcal{N}(\mu(\varphi_t), \Sigma(\varphi_t)) \| \mathcal{N}(0, I))$
\end{center}

The Transformer inputs $\hat{c}_t$ are first added to positional encodings to properly influence transduction, and then context-aware conditioning vectors $\overline{c}_t$ are produced from the position encoded inputs. The context-informed vectors $\overline{c}_t$ are concatenated with Gaussian noise $z_t\sim p_z$, where $p_z$ is the random input prior $z\sim\mathcal{N}(0,1)$. This combined input is fed through a fully connected (FC) layer, mapping each instance to dimension $C\times H\times W$, where H, W are the height and width of the initial image channels to be upsampled, and C their channel number. This output mapping is rearranged in a tensor $I_t\in\mathbb{R}^ {C\times H\times W}$ and fed through a set of residual upsampling blocks, similar to \cite{zhang2019sagan}. The purpose of a residual block \cite{he2016resnet} is to learn a mapping $F(x)=H(x)-x$ where $H(x)$ is the actual desired mapping in the underlying distribution. The final output is produced utilizing a skip connection such that $\hat{H}(x) = F(x)+x$. 
%Learning the residual is easier than learning the original transformation.
%and prove successful in countering the accuracy degradation observed with increasing network depth.
In each upsampling block, the input image features \textbf{$I_t$} are normalized via Batch Normalization \cite{ioffe2015batchnorm} and passed
through a \textit{ReLU} activation. Then, both spatial dimensions are doubled via nearest-neighbor upsampling,
and a convolutional filter is applied to transform image features, while halving the channel dimension to mitigate computational complexity as the image planes get larger. The tensor is again normalized and passed
through a \textit{ReLU} activation as well as a final convolutional filter. In order to match the spatial input and output dimensions we perform a minimal transform on the skip connection, using
nearest-neighbor upsampling and passing through a learned $1 \times 1$ convolutional filter.
After feature upsampling to the desired dimension $H \times W$, a final $3 \times 3$ convolution layer is used to
produce a 3-channel image, followed by a \textit{tanh} activation to remap pixel values into
[$-1, 1$]. We also use Spectral Normalization to further stabilize the training process. The entire image sequence can be generated in parallel, greatly improving training efficiency. 

\begin{figure}[t]
\hspace{-0.2cm}
\includegraphics[width=9cm]{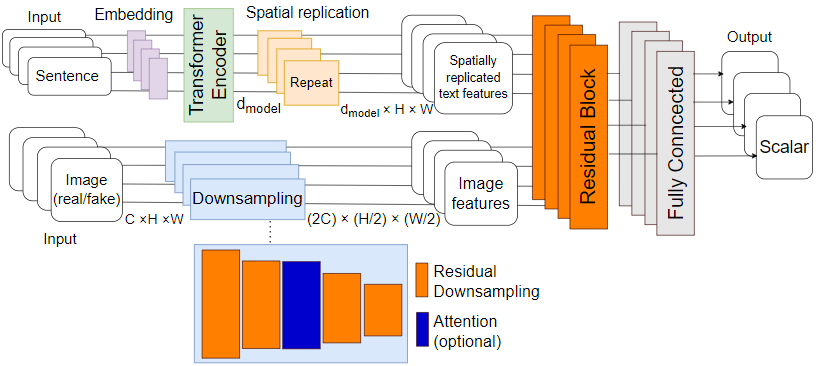}
\caption{Image discriminator $D_{im}$ (T = 4 frames)}
\vspace{-0.2cm}
\label{fig:im_dscr}
\end{figure}

\subsection{Image Discriminator}
The image discriminator $D_{im}$ (Fig. \ref{fig:im_dscr}) is tasked to discern between real and generated images individually. To that end, $D_{im}$ utilizes the input features $\phi_t$ of each individual sentence corresponding to a story frame, the context, and the image $I_t$ itself to be evaluated. The context is important for $D_{im}$, because each frame in a story depends on the rest to form many of its details. 
Each image to be evaluated is passed through a series of residual downsampling blocks. 
Image features from each layer are first passed through a \textit{Leaky ReLU}, then from a spectrally normalized convolutional layer, remapping the $C \times H \times W $ tensor to double the channels. 
After another \textit{Leaky ReLU}, a spectrally normalized strided convolution layer downsamples the image features. 
We prefer this option over a pooling layer due to the inferences made by Radford et. al in \cite{radford2016unsupervised}. 
%The final tensor has dimension $(2C) \times (H/2) \times (W/2)$. 
All images are evaluated in a batch to take advantage of the Transformer's parallel processing. Dropout in all $D_{im}$ residual blocks is proven beneficial, 
to prevent overfitting and overt coupling of individual layer units. 
%We attempt the usage of intra-image attention in order to assist the network in learning longer-range relationships of image features. 
To produce an output scalar, each vector of dimension $d_{model}$ given by the encoder is spatially replicated to create a 
$d_{model} \times H \times W$ tensor that is then concatenated with the image features along the channel axis.  
These features are passed through a residual block to jointly learn from image and text features. A final FC layer mapping features to a single scalar leads to a sigmoid activation function, ultimately producing a probability $D_{im}(I_t) \in [0, 1]$.

\subsection{Story Discriminator}
\begin{figure}[t]
\centering
\includegraphics[width=8.3cm]{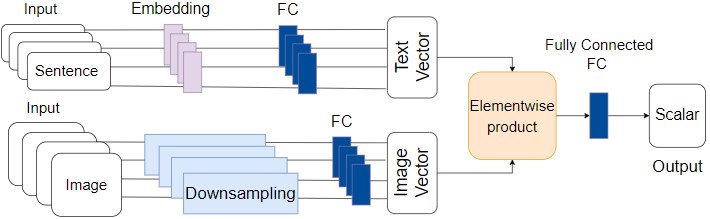}
\caption{Story discriminator $D_{st}$ (T = 4 frames)}
\label{fig:st_dscr}
\end{figure}

The story discriminator $D_{st}$ (Fig. \ref{fig:st_dscr}) enforces consistency and meaningful progression along the image sequence $I=(I_1, ... , I_T)$ by jointly learning a common feature space for text and images. The image features are downsampled using similar residual blocks as in $D_{im}$. All image features for the same story are concatenated into a single storyboard vector. On the text side, a FC layer maps all sentence embeddings $\textbf{\textit{S}}=(\phi_1, . . . , \phi_T)$ to vectors in this shared space, also concatenated into one big text feature vector. The two story-wide vectors are then multiplied
elementwise and the result is passed through a FC layer to output a scalar similarity score $D_{st}$.

\subsection{Training}
Training requires minimizing $\mathcal{L}_{im},\mathcal{L}_{st}, \mathcal{L}_G$: \hfill
\vspace{-0.3cm}
\begin{multline*} 
\hspace{-0.3cm}
\mathcal{L}_{im} =
\sum_{t=1}^T ( 
\mathbb{E}_{(i_t, \varphi_t)} [logD_{im}(i_t, \varphi_t, h_0; \psi_I)] + \\
\mathbb{E}_{(z_t, \varphi_t)} [log(1-D_{im}(G(z_t, \varphi_t;\theta), \varphi_t, h_0; \psi_I))] ), \\
\hspace{-3.7cm}
\mathcal{L}_{st} =
\mathbb{E}_{(\mathbf{I}, \mathbf{S})} [logD_{st}(\mathbf{I}, \mathbf{S}; \psi_S)] + \\
\mathbb{E}_{\epsilon, \mathbf{S}} [log(1-D_{st}([G(z_t, \varphi_t; \theta)]_{t=1}^T), \mathbf{S};\psi_S))], \\
\hspace{-1.0cm}
\mathcal{L}_G = 
\mathbb{E}_{(z_t, \varphi_t)} [log(D_{im}(G(z_t, \varphi_t;\theta), \varphi_t, h_0; \psi_I))] + \\
\mathbb{E}_{\epsilon, \mathbf{S}} [log(D_{st}([G(z_t, \varphi_t; \theta)]_{t=1}^T), \mathbf{S};\psi_S))] +
Loss_{KL}    
\end{multline*}
where $z_t \sim p_z$, and $h_0$ serves as story embedding. The alternative formulation following \cite{goodfellow2014gans} is employed for $G$ to provide sufficient gradients. We also
use the matching aware discriminator criterion as in \cite{ref2}. One-sided label smoothing is utilized by setting positive labels to 0.9 instead of 1.0 to avoid the pitfalls of regular label smoothing \cite{ref4}.

%and we experiment with different learning rate values and scheduling schemes for the three networks, in consideration of the two time-scale update rule \cite{heusel2018ttur}, while maintaining balanced updates. 

\section{Experiments}
\label{sec:experiments}
We present results on CLEVR-SV \cite{clevr}, focusing on cases where objects may not be clearly separated or even occluded. This issue, despite its significance, was not addressed in prior work. For all experiments, Adam optimizer \cite{adam} is used for gradient descent with $\beta_1$ = 0.5 and $\beta_2$ = 0.999. After extensive hyperparameter tuning we present results on the original Transformer with $d_{model}$ = 512, $N_{heads}$ = 8, $N_{layers}$ = 6. 
%Reducing the number of heads proved to be immediately detrimental to performance, while wider or deeper transformers proved to have too much representational capacity.
\subsection{Impartial Transformer Encoder}
%Transformer-encoded context-imbued vectors are necessary for $G$ and $D_{im}$, while we consider that $D_{st}$ is able to learn sufficient mappings for conditional embeddings and images jointly without any immediate need for other processing, as it considers entire sequences in parallel.
We explore the option of utilizing one \textbf{\textit{Impartial transformer}}  encoder, whose parameters are updated jointly
by $G$ and $D_{im}$. We hypothesize such an encoder would learn a task-conducive representation for embedding sequences by simply encoding necessary context without giving an advantage to either adversary. We further attempted to train the encoder to also receive gradients from the $D_{st}$, but found this addition to be confusing the encoder, to the point of learning completely mismatched representations of the context space. 
\subsection{Learning rate schemes}
Motivated by the Two Time-scale Update Rule \cite{heusel2018ttur}, we attempt to find an optimal learning rate scheme for the three networks while maintaining a 1/1/1 update ratio for more efficient training, thus proposing a \textbf{\textit{Three Time-scale Update Rule}}. 
%The architecture we use for these tests is shown in Fig. \ref{fig:gen}, \ref{fig:im_dscr}, \ref{fig:st_dscr} without any attention in the image scaling layers, and separate transformer encoders for $G$ and $D_{im}$.  
After 20 epochs, the learning rates are halved based on a typical scheduling scheme.  We observe that when $G$ learns faster than the discriminators, the whole model suffers from mode
collapse: $G$ easily fools both discriminators early on, leading training to a stalemate since the
discriminators cannot produce any meaningful gradients to guide generation. When maintaining a low learning rate for $G$, increasing the $D_{im}$ learning rate proves to lead $G$ into creating images that correspond better to the conditioning. $G$ is faster in learning the correct matching for color and shape between image and description
vector, as well as learning to produce more concrete shape features, at least for large objects.
When increasing the learning rate of $D_{st}$, we immediately observe greater consistency across
images. 
%Objects maintain their position throughout the story and the correct number of objects for each image is generated more often than not. 
Lower learning rates also seem to affect text-image matching, with
$G$ creating images with wrong color, shape and size more frequently.
We thus argue that it is beneficial for the two discriminators to learn about 4 times as fast as
$G$. Specifically, we find $lr_G$ = 0.0001, ${lr_D}_{im}$ = 0.0004, ${lr_D}_{st}$ = 0.0004 to be optimal, as higher learning rates proved to be too fast for convergence. 

\begin{figure*}[h!]
    \centering
    \begin{subfigure}{0.33\textwidth}
    \includegraphics[width=0.95\columnwidth]{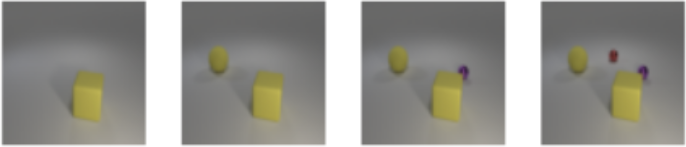}
    \end{subfigure}
    \begin{subfigure}{0.33\textwidth}
     \includegraphics[width=0.95\columnwidth]{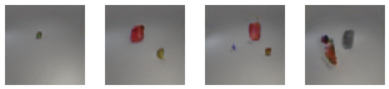}
    \end{subfigure}
    \begin{subfigure}{0.33\textwidth}
    \includegraphics[width=0.95\columnwidth]{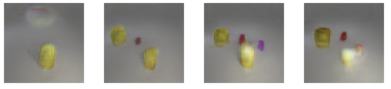}
    \end{subfigure}
    \caption*{(a) Left: Ground truth (T=4). Middle: StoryGAN generated frames, low relevance and object quality. Right: Ours, baseline.}
    \centering
    \begin{subfigure}{0.33\textwidth}
    \includegraphics[width=0.95\columnwidth]{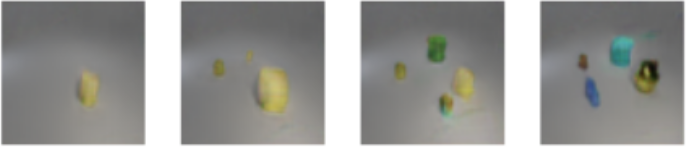}
    \end{subfigure}
    \begin{subfigure}{0.33\textwidth}
    \includegraphics[width=0.95\columnwidth]{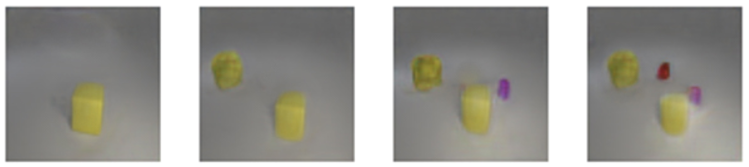}
    \end{subfigure}
    \begin{subfigure}{0.33\textwidth}
    \includegraphics[width=0.95\columnwidth]{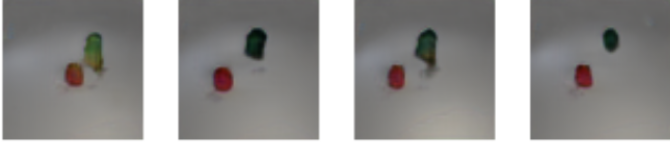}
    \end{subfigure}
    \caption*{(b) Our results without attention. Left: Separate Transformer Encoder for $G$, $D_{im}$, $D_{st}$, low object relevance. Middle: Impartial Encoder ($G$ and $D_{im}$ gradients). Right: Impartial encoder (all $G$, $D_{im}$, $D_{st}$ gradients), mode collapse.}
    \caption{Ablation studies of our framework indicate the power of the Impartial Transformer ($G$ and $D_{im}$ gradients).}
\label{fig:ablations}
\end{figure*}

\begin{figure*}[h!]
    \centering
    \begin{subfigure}{0.33\textwidth}
        \includegraphics[width=0.96\columnwidth]{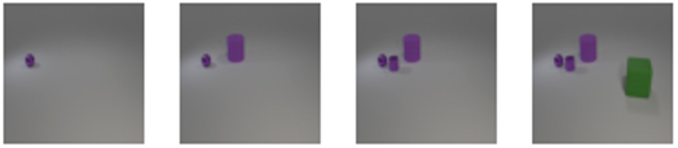}
        %\caption{Ground truth story frames with occlusion}
    \end{subfigure}
    \begin{subfigure}{0.33\textwidth}
        \includegraphics[width=0.96\columnwidth]{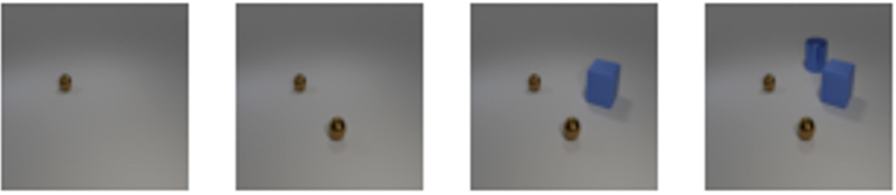}
        %\caption{Separate Transformer Encoder for $G$, $D_{im}$, $D_{st}$}
    \end{subfigure}
    \begin{subfigure}{0.33\textwidth}
        \includegraphics[width=0.96\columnwidth]{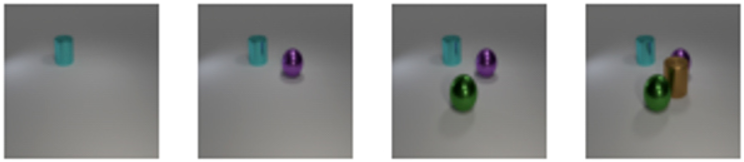}
        %\caption{Impartial Encoder ($G$ and $D_{im}$ gradients)}
    \end{subfigure}
    \begin{subfigure}{0.33\textwidth}
        \includegraphics[width=0.96\columnwidth]{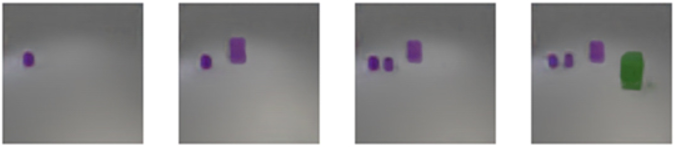}
        %\caption{Impartial Encoder (All $G$, $D_{im}$, $D_{st}$ gradients)}
    \end{subfigure}
        \begin{subfigure}{0.33\textwidth}
        \includegraphics[width=0.96\columnwidth]{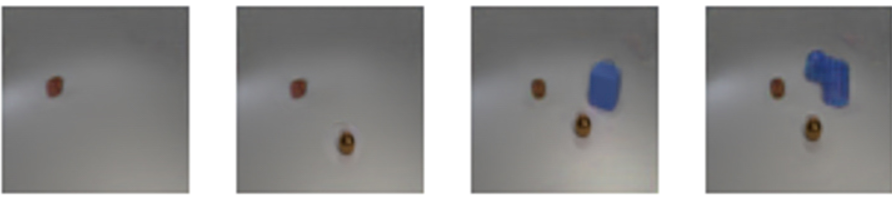}
        %\caption{Impartial Encoder ($G$ and $D_{im}$ gradients)}
    \end{subfigure}
    \begin{subfigure}{0.33\textwidth}
        \includegraphics[width=0.96\columnwidth]{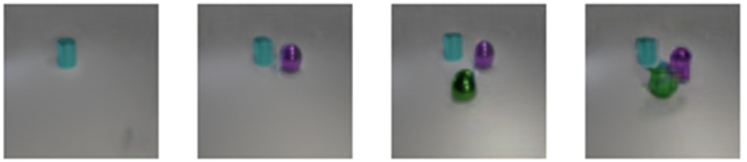}
        %\caption{Impartial Encoder (All $G$, $D_{im}$, $D_{st}$ gradients)}
    \end{subfigure}
      \begin{subfigure}{0.33\textwidth}
        \includegraphics[width=0.97\columnwidth]{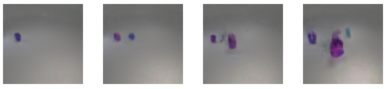}
        %\caption{Impartial Encoder (All $G$, $D_{im}$, $D_{st}$ gradients)}
    \end{subfigure}
        \begin{subfigure}{0.33\textwidth}
        \includegraphics[width=0.97\columnwidth]{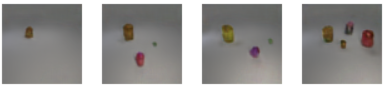}
        %\caption{Impartial Encoder ($G$ and $D_{im}$ gradients)}
    \end{subfigure}
    \begin{subfigure}{0.33\textwidth}
        \includegraphics[width=0.96\columnwidth]{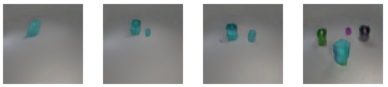}
        %\caption{Impartial Encoder (All $G$, $D_{im}$, $D_{st}$ gradients)}
    \end{subfigure}
    \caption{(a) 1st row ground truth, (b) 2nd row generated frames (ours-Impartial Transformer), (c) 3rd row generated frames (storyGAN) of 3 stories with T=4. From left to right (every 4 images) difficulty of stories increases due to object occlusion.}
    \label{fig:examples}
\end{figure*}

\subsection{Warmup Scheduler}
We  experiment with decaying the learning rate by halving it every 20 epochs.
The original Transformer \cite{transformer} recommends a specific learning rate scheduling scheme to be used along with the Adam optimizer: The learning rate should first be increased linearly for a number of warmup
steps and then decreased proportionally to the inverse square root of the number of total steps, where one step is considered to be a single batch of data passing through the network. We observe that the scheduler fails to train the context encoder, resulting in mostly nonsensical representations. We presume this is because the recommended optimizer only takes into account $d_{model}$ and the number of warmup steps, forcing the learning rate to generally remain much
higher than what the learning rates of the Adam optimizer in regular decay are,  preventing network
from convergence.

\subsection{Results}
Visual results including ablations are presented in Fig \ref{fig:ablations}, while comparison over easy and hard examples are presented in Fig. \ref{fig:examples}. There is an obvious improvement over StoryGAN \cite{storygan}, which fails to generate the proper sequence, and also lacks in fidelity. The second row of Fig. \ref{fig:ablations} indicates the optimal usage of the \textbf{\textit{Impartial transformer}}.
Even though our implementation presents satisfactory results when objects are placed in a distance from each other (Fig \ref{fig:examples}, left), in cases when objects are adjacent or overlap, there are some sacrifices to be made: either semantics -especially shape and material- are not distinct enough (Fig \ref{fig:examples}, middle), or objects are 'swallowed' by their neighbors (Fig \ref{fig:examples}, right), which results in low quality semantics. 
%Our framework promotes consistency and relevance, even in cases of occlusion, which sometimes may result in blurred object shapes or color sharing. This highlights a trade-off between visual quality vs consistency and relevance over hard examples generation. 
The results of human evaluation experiments over preference are presented in Table \ref{tab:human_eval}. Results using automated metrics are presented in Table \ref{tab:global}. Our framework clearly outperforms prior efforts \cite{storygan, Maharana2021ImprovingGA, Maharana2021IntegratingVL} according to Clean-FID \cite{cleanfid}, LPIPS \cite{lpips} and SSIM. We mainly focus on LPIPS metric for comparison that reflects human perception, where we achieve 16\% improvement over prior approaches \cite{storygan, Maharana2021ImprovingGA, Maharana2021IntegratingVL}.

\begin{table}[htp]
\centering
\caption{Human Evaluation preference (averaged results), Win\% = \% times our output stories were preferred
over \cite{storygan}, Lose\% for vice-versa, Tie\% when equally preferred.}
\label{tab:human_eval}
  \begin{tabular}{cccc}
    \toprule
    Attribute & Win\% & Loose\% & Tie\%\\
    \midrule
    Visual Quality & 25& 20 & 55\\
    Consistency & 37& 32& 31 \\
    Relevance  & 32& 30& 38\\
  \bottomrule
\end{tabular}
\end{table}

\begin{table}[t!]
\caption{Average evaluation metrics. }
\label{tab:global}
\begin{tabular}{p{20pt}|p{57pt}p{42pt}p{50pt}p{20pt}}
\toprule
Frame & FID$\downarrow$ & Clean-FID$\downarrow$ & LPIPS$\downarrow$ & SSIM$\uparrow$  \\
\midrule
1st & 32.94 $\pm$ 7.85 & 111.20  & 0.18 $\pm$ 0.06 & 0.81  \\
2nd & 37.41 $\pm$ 6.67 & 110.80  & 0.19 $\pm$ 0.05 & 0.73  \\
3rd & 47.41 $\pm$ 15.83 & 106.69  & 0.23 $\pm$ 0.05 & 0.68\\
4th  & 48.41 $\pm$ 3.84 & 133.15  & 0.25 $\pm$ 0.05 & 0.62  \\
\midrule
All & 41.54 $\pm$ 8.55 & \textbf{115.46} & \textbf{0.21} $\pm$ 0.05 & \textbf{0.71}  \\
\cite{storygan} & 41.45 $\pm$ 6.25 & 123.40 & 0.25 $\pm$ 0.03 & 0.65\\
\cite{Maharana2021IntegratingVL} & 41.96 $\pm$ 9.66 &  124.97 &  0.25 $\pm$ 0.08 &  0.67 \\
\cite{Maharana2021ImprovingGA}  & 41.80 $\pm$ 8.81 &  122.62 &  0.25 $\pm$ 0.05 &  0.68 \\
\bottomrule
\end{tabular}
\footnotesize{'All' refers to global results of the Impartial Transformer and is compare with the global results of \cite{storygan}, \cite{Maharana2021IntegratingVL}, \cite{Maharana2021ImprovingGA}. Results from \cite{Maharana2021IntegratingVL}, \cite{Maharana2021ImprovingGA} are obtained by re-training on CLEVR-SV.}
\end{table}

\section{Conclusion}
In this work, we developed a transformer-inspired framework for story visualization, aiming to set a new baseline in literature by achieving improvements according to perceptual metrics. The usage of the \textit{\textbf{Impartial Transformer}} demonstrated promising directions for the evolution of generative models in the same track, as few -if any- current implementations exploit a 'forking' module jointly trained by two adversaries. As future work we plan to explore the evaluation part of SV.

\bibliographystyle{IEEEbib.bst}
\bibliography{strings,refs}

\end{document}